\documentclass[journal]{IEEEtran}
\IEEEoverridecommandlockouts
\usepackage{amsmath,amssymb,amsfonts}
\usepackage{algorithmic}
\usepackage{graphicx}
\usepackage{textcomp}
\usepackage{xcolor}
\usepackage{geometry}
\usepackage{cite}
\usepackage{array}
\usepackage{mdwmath}
\usepackage{mdwtab}
\usepackage{eqparbox}
\usepackage{url}
\usepackage{graphics} 
\usepackage{epsfig} 
\usepackage{mathptmx} 
\usepackage{times} 
\usepackage{amsmath} 
\usepackage{amssymb}  
\usepackage{booktabs} 
\usepackage{algorithm}
\usepackage[justification=centering]{caption}
\usepackage{gensymb}
\def\BibTeX{{\rm B\kern-.05em{\sc i\kern-.025em b}\kern-.08em
    T\kern-.1667em\lower.7ex\hbox{E}\kern-.125emX}}
\def\BibTeX{{\rm B\kern-.05em{\sc i\kern-.025em b}\kern-.08em
    T\kern-.1667em\lower.7ex\hbox{E}\kern-.125emX}}

\begin{document}

\title{An Embedded Iris Recognition System Optimization using Dynamically Reconfigurable Decoder with LDPC Codes\\
}


\author{Longyu Ma,
        Chiu-Wing Sham,~\IEEEmembership{Senior Member,~IEEE,}
        Chun Yan Lo,
        and~Xinchao Zhong
  \thanks{The authors are with the School of Computer Science, The University of Auckland, Auckland 1142, New Zealand (e-mail: lma792@aucklanduni.ac.nz; b.sham@auckland.ac.nz; clo871@aucklanduni.ac.nz; xzho323@aucklanduni.ac.nz).}}

\maketitle
\begin{abstract}
Extracting and analyzing iris textures for biometric recognition has been extensively studied. As the transition of iris recognition from lab technology to nation-scale applications, most systems are facing high complexity in either time or space, leading to unfitness for embedded devices. In this paper, the proposed design includes a minimal set of computer vision modules and multi-mode QC-LDPC decoder which can alleviate variability and noise caused by iris acquisition and follow-up process. Several classes of QC-LDPC code from IEEE 802.16 are tested for the validity of accuracy improvement. Some of the codes mentioned above are used for further QC-LDPC decoder quantization, validation and comparison to each other. We show that we can apply Dynamic Partial Reconfiguration technology to implement the multi-mode QC-LDPC decoder for the iris recognition system. The results show that the implementation is power-efficient and good for edge applications.
\end{abstract}


Biometric recognition has come into being a vital research field in today's networked society. It is a measurement of person's biological characteristic for the purpose of determining human identity. One of its advantage is high-degree assurance of the individual's identity, which leads to safer personal privacy than other feasible but vulnerable solutions, such as passwords or Personal Identification Numbers (PINs). Iris, as the most promising biometrics in terms of reliability and security~\cite{bodade2014iris, masek2003recognition}, is used for authentication and security purposes. Nevertheless, intrinsic problems of any biometric, including but not limited to iris, fingerprints, face, are unavoidable. One of these is that some small variations between reference iris data and probe iris data simply cause a wide divergence of outcomes. One~\cite{tan2004iris} attempts to present an overview of recent progress in iris recognition and remaining challenges, which is image acquisition, liveness detection, iris normalization, feature extraction and matching. With different image processing methods, such a defect may be mitigated, but most of them need supports from OpenCV libraries huge Neuron Networks~\cite{vspetlik2019iris} or still in a preliminary stage, resulting in lacking hardware-friendly approaches. With the advancement of integrated circuits technology such as placement\cite{1688966},  routing~\cite{6523615}, clock planning~\cite{LU2012121} and buffer planning~\cite{998375}, biometric recognition can be applied in a wider range of edge applications\cite{4375772}.

Apart from the methods mentioned above, researchers like \cite{kanade2009cancelable} introduce a novel way to use error correcting codes (ECC) to improve iris recognition systems. Then LDPC codes start to attract their attention. Low-density parity-check (LDPC) codes~\cite{gallager1962low, 681360} are one of error correction codes which can be used to handle the errors in an error-prone communication channel. There have been comprehensive studies~\cite{7932182, 9179021, 9215273} on improving the complexity and accuracy of the LDPC decoders in communication fields such as applying appropriate calculations~\cite{6419075, 7388304, 8587568, 9115068} and different structure of LDPC codes~\cite{6481477, 7370816, 8629018}.

In additional to communications, LDPC code decoders have bid their way into other areas such as storage~\cite{GCCE201902, 7804048, 8425687} or biometric systems~\cite{baldi2011fuzzy, uludag2004biometric, sutcu2008feature, slepian1973noiseless}. Previous studies have shown that certain error correction codes can increase the performance of biometric systems. One novel iris recognition system using LDPC and SHA~\cite{seetharaman2012ldpc} is proposed, in which paper an LDPC encoder and decoder is connected to the module for the purpose of error-correction, but only simulation results are shown. ~\cite{5489230,moi2017modified} expend their experiments on four different iris database and confirm the contribution of ECC or LDPC to the biometric field. Being one of the most crucial module, an implementable LDPC decoder should be considered, especially for embedded devices. 

The Quasi-Cyclic LDPC codes, as the subclass of LDPC codes are more hardware-friendly, easier to be applied in parallel. Thanks to the widespread exploitation of LDPC codes since decades ago, QC-LDPC codes are proved to be one of the most promising candidates for having their structures optimized for hardware\cite{4114369}. However, there are not many studies exploring the implementation of QC-LDPC codes into biometric systems~\cite{8977863}. emphasizes how to build a mechanism to exchange information between a processor and QC-LDPC decoders and shows the utilization of each modules in its design, while calculates the Hamming Distance between LDPC-based and non-LDPC-based iris recognition in a fixed threshold~\cite{9081255}. Both of them lack a thorough performance analysis and multidimensional consideration of the QC-LDPC feasibility in a biometric system.   
\begin{figure*}[htbp]
\centerline{\includegraphics[width=1\textwidth]{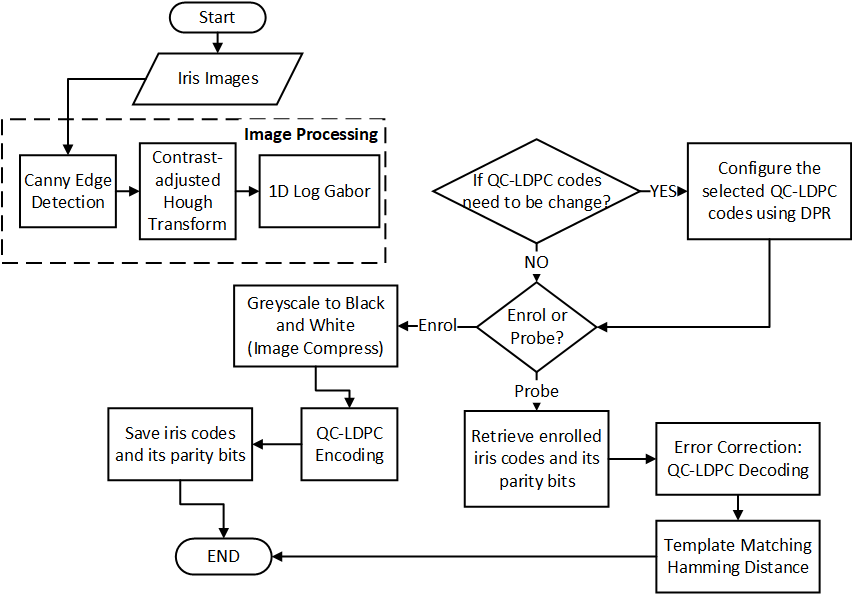}}
\caption{Flow Chart of Proposed System}
\label{fig40}
\end{figure*}
To alleviate the drawback mentioned above, we implement an iris recognition system onto an embedded device with dynamic multiple QC-LDPC decoders to improve the recognition accuracy. The main points of this paper are:

\begin{itemize}
  \item On the basis of performance and complexity analysis, proper QC-LDPC codes are selected from IEEE 802.16 and are implemented in a format of different quantized fix-points(2 bits, 4 bits, 8 bits).  
  \item A basic iris recognition system is kept as a program in Linux setting up an inner-chip link with programmable logic filled with QC-LDPC decoders which are configured dynamically in a circuit level using FPGA Dynamic Partial Reconfiguration (DPR). 
  \item Encoding and decoding iris data before matching are experimentally proved to improve the true acceptance rate on the iris database(CASIA-IrisV4-Interval).
\end{itemize}

In section II, the proposed architecture with graphics is presented. Then, the performance evaluation experiments are described in the third part of the paper. The paper comes to a logical conclusion in the last section.

\section{Proposed Architecture}

The proposed architecture consists of two major sections. One is a minimal version of the iris image process: its algorithms in this section stems from University of Salzburg Iris-Toolkit v3.0\cite{USIT3}, comprising image segmentation, localization, feature extraction in Fig.\ref{fig1}. The other section is based on the previous section and it mainly contains QC-LDPC decoders.

\subsection{Minimal Iris Image Processing Units Establishment}

The proposed implementable optimization for iris recognition is based on conventional iris image processing units. In order to present the following sections, these units should be specified briefly.

Generally, an iris image are supposed to be acquired, boundary-detected, segmented, localized and feature-extracted. The minimal iris image processing units collect images from CASIA-Internal V4.0 database. Then Canny Edge Detection, Contrast-adjusted Hough Transform is adopted in segmentation process. The next process is normalization using Doughman rubber sheet model. At last, 1D LogGabor~\cite{rathgeb2012iris,kahlil2010generation}, as a method of feature extraction, will be performed before iris codes are pop out. The iris codes, in the proposed design, subsequently are fed into QC-LDPC encoders or decoders. As they are developed based on OpenCV Library, they are kept in an ARM-based Linux depicted in the top-left of Fig.\ref{fig1}.   

\subsection{Error Correction Procedure}

\begin{figure*}[htbp]
  \centering
  \includegraphics[width=1\textwidth]{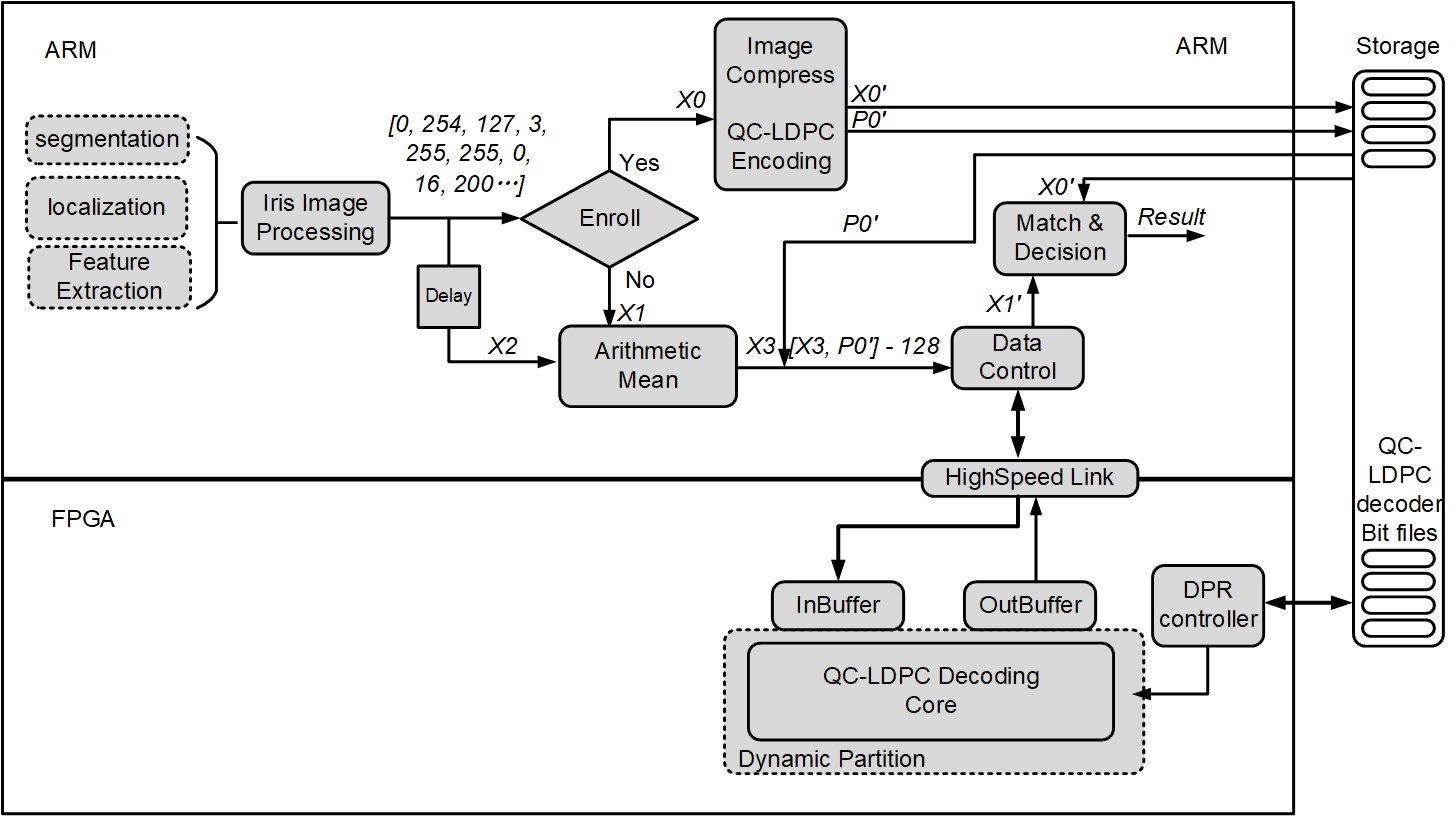}
  \caption{Block Diagram For Proposed Design}
  \label{fig1}
\end{figure*}

A thorough procedure of iris recognition should be twofold. One is called enrolment phase, during which a reference iris image is recorded in the system. The other is probe or verification phase, during which new irises are acquired and compared to the previous one. Under a known threshold, the final result would be either match or not match. Our proposed QC-LDPC-code-related modules are involved in both two fundamental phases. Basic blocks denoted as major modules are depicted in Fig. \ref{fig1} which consists of 7 stages. The system starts with the image processing stage, including segmentation, localization, feature extraction, DPR-based QC-LDPC code selection, QC-LDPC encoding, QC-LDPC decoding and template matching.

\begin{enumerate}

    \item  Enrolment Phase
    \begin{enumerate}
        \item  A new block of iris codes from Iris Image Processing in Fig. \ref{fig1} will be decided as binary digits. The new block is denoted as $X0$.
        \item  The QC-LDPC encoder produces a new block of codeword by encoding the compressed $X0'$ and output parity bits $P0'$ (the length of $P0'$ is $n$).
        \item $P0'$ would be stored in an external storage, like RAM or FLASH.
    \end{enumerate}
    \item Verification Phase
        \begin{enumerate}
        \item Another iris image from an intruder or genuine person is captured and processed until $X1$ is produced.
        \item Same procedures will be done again to generate $X2$ and its length is $n$. Subsequently, $X1$ and $X2$ are both sent to Arithmetic Mean module.
        \item Arithmetic Mean is added to increase the acceptance rate. $X1$ and $X2$ come from the same iris but are captured in different time slots. The output of Arithmetic Mean, as known as $X3$, is a new vector of iris codes with the same number of elements as $X1$ and $X2$, where each element is the average of the elements in the same location of $X1$ and $X2$. The output of Arithmetic Mean module will be transferred to the QC-LDPC decoder or Matching Module directly.
        \item Concatenated with bit-extended $P0'$, this new group of coded data($[X3,P0']$) will be fed into the QC-LDPC decoding module. Please be noted that a unsigned-to-signed operation (subtracting 128) will be done before QC-LDPC decoding starts.
        \item The decoded vector $X1'$, as the output of QC-LDPC decoder will be sent back to ARM-side followed by $X0'$ for matching.
        \item In the matching module, the main purpose is to find the minimum value, know as Hamming Distance(HD). Two iris samples, presented as iris binary codes, are input data of matching modules. After comparing one's HD to the other, a decision is given based on current matching threshold.
    \end{enumerate}
\end{enumerate}

While Fig. \ref{fig1} demonstrates the proposed design module by module, Fig. \ref{fig40} presents the detailed flow chart of proposed system, listing where iris data should go and be processed in order. 

\begin{figure*}[htbp]
  \centering
  \includegraphics[width=1\textwidth]{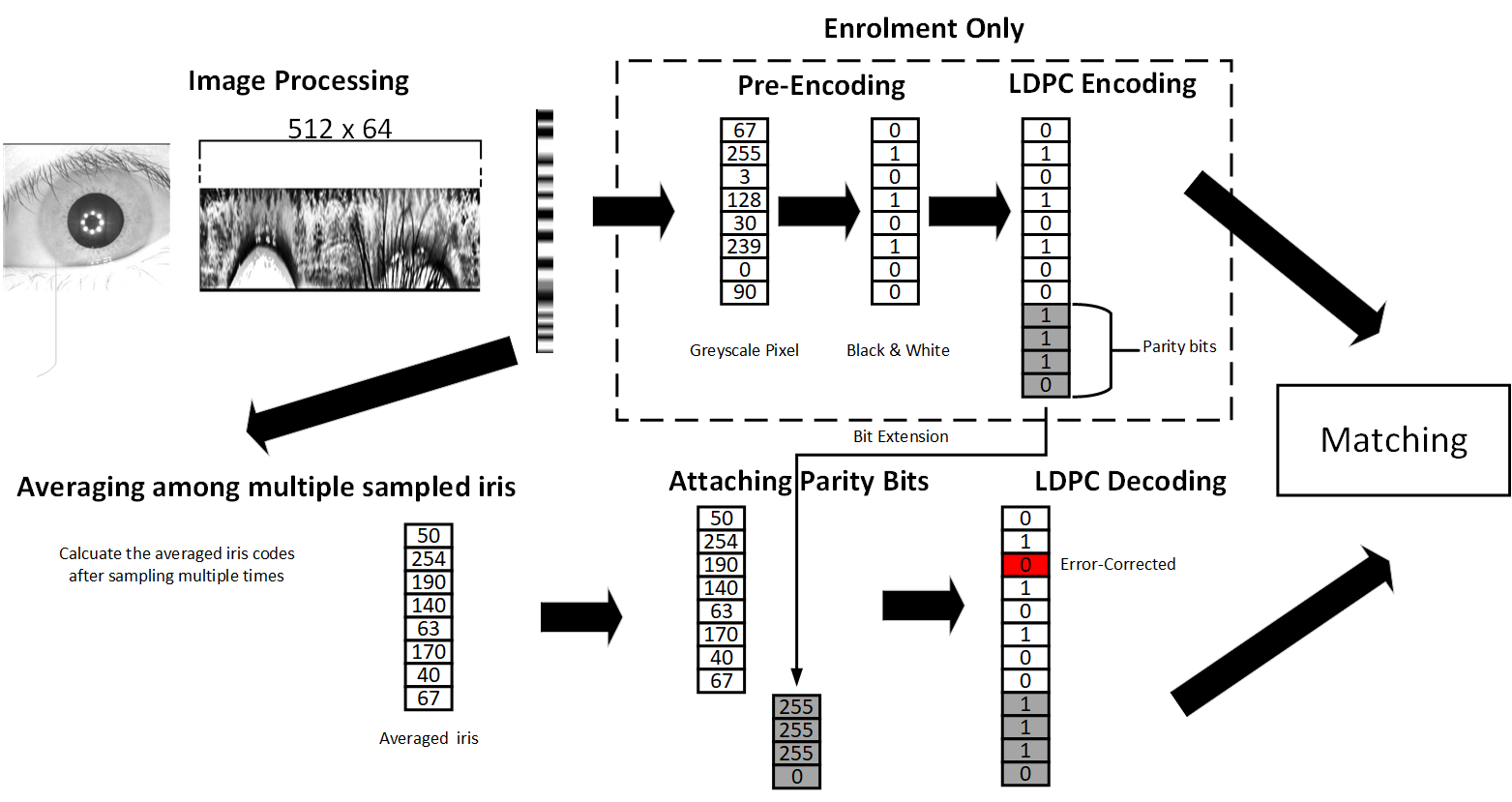}
  \caption{Proposed LDPC-related error-correcting method}
  \label{fig100}
\end{figure*}

Fig. \ref{fig100} shows the overall process of the proposed QC-LDPC error-correcting method. Briefly speaking, parity bits of the reference iris are used to correct the pre-verified irises which are multiple-sampled and then averaged. After LDPC decoding, the outputs are the hard-decision values and the highlighted value has been error-corrected.

\subsection{Log Likelihood Ratio Soft-decision QC-LDPC Decoder}

A previous piece of work~\cite{seetharaman2012novel} adopts an LDPC scheme rather than QC-LDPC and it feeds LDPC decoders with hard decision. Some information may lose when multi-bit data are compressed into binaries. In this paper, quantized iris code will be maintained to feed the QC-LDPC decoder during probe phase. One of the advantages is that a soft-decision decoder has been known to typically perform better than a hard-decision decoder~\cite{haykin1988digital} and can provide
1–2 dB of additional net coding gain. Not like hard-decision decoders, which need a series of binaries, the inputs to a soft-decision decoder may take on a whole range of values in-between. 

The architecture of QC-LDPC decoder in this paper origins in \cite{6419075}. As for QC-LDPC decoding architecture, a layered decoding structure\cite{8587568} is adopted to improve parallelism. The number of Check Node Processors and Variable Node Processors is adjusted based on QC-LDPC H matrix in IEEE 802.16.  

\subsection{FPGA Dynamic Partial Reconfiguration and Multiple Decoders}
FPGA Dynamic Partial Reconfiguration derives from \cite{8906146}. It provides a circuit-level real-time flexible decoder architecture. Those one-LDPC-code-supportive QC-LDPC decoders, using DPR, can be time-multiplexed and integrated, acting as a multiple QC-LDPC codes decoder. The merit of this technology is that it can hold other circuits in the FPGA untouched. When the new circuit programmed by the corresponding bit file, the whole FPGA will continue running its logic. The dynamic partition of circuit of FPGA must be specified, otherwise, a tradition configuration need to be do, cause everything in the FPGA reset. 

The reason why the proposed system adopts DPR is one class of QC-LDPC codes in IEEE 802.16 sometime is not enough to fulfill all scenarios. For example, when LDPC decoder algorithm is changed from floating-point version(simulation version) to fix-point version(implementation), Algorithm with few quantization bits usually consumes less power, less complex, but perform less accuracy. This configuration may be not good in simulation, but feasible in some resource-constraint devices. Moreover, the difference in QC-LDPC codes may also influence the system. If DPR is adopted, a real-time configuration can happen in a ten-millisecond level.    

\begin{figure*}[htbp]
\centerline{\includegraphics[width=1\textwidth]{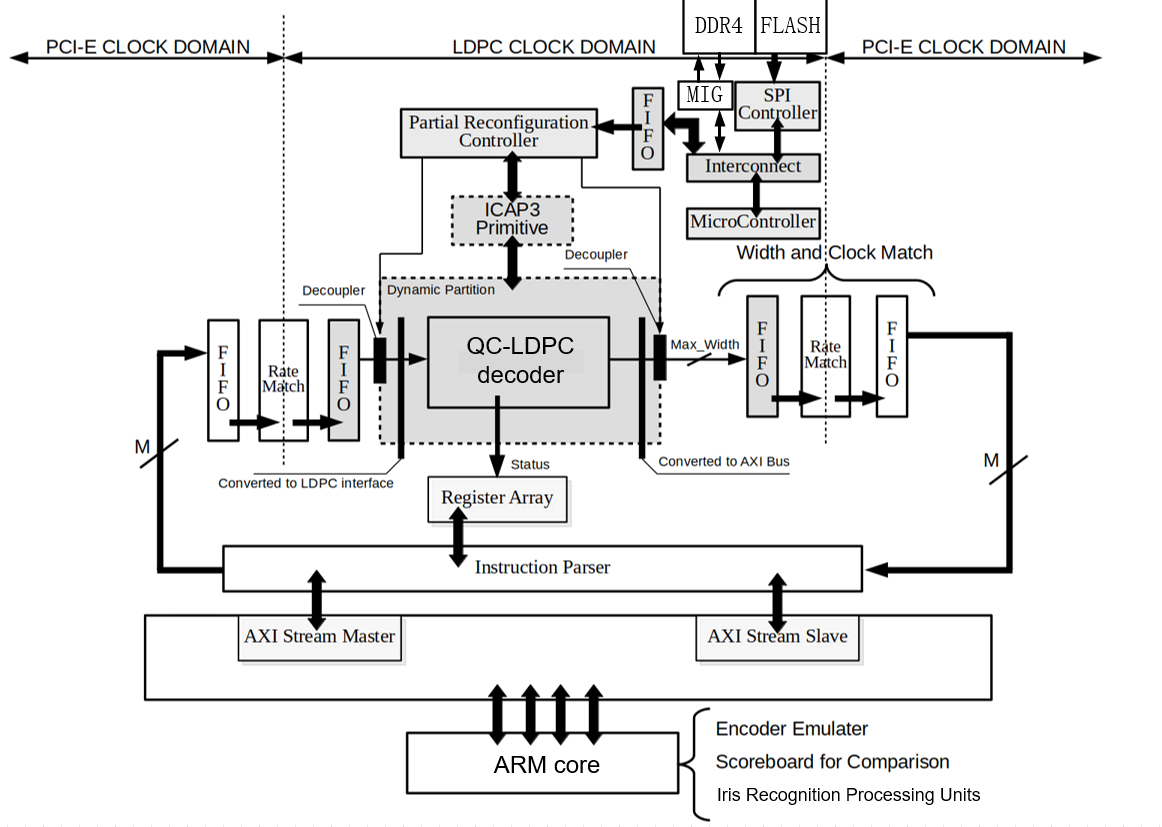}}
\caption{The implementation of the proposed iris recognition system using multiple QC-LDPC codes}
\label{fig4}
\end{figure*}

Fig.\ref{fig4} demonstrates the DPR-related components and relationship with ARM cores which include iris recognition processing units. It is an implementation on the basis of Fig. \ref{fig1}.  

Basically, iris codes are sent from ARM cores to a QC-LDPC decoder and the results will be delivered to ARM cores for follow-up process through AXI buses. Around the decoder, FIFOs and Rate Matches acting like buffers to receive data temporarily before move data to AXI buses. The DPR controller module can change the circuit in the DPR Partition on the fly and that part of circuit represents one selected QC-LDPC decoder. As soon as they confirm a request from ARM cores to change a QC-LDPC decoder, they will shut down the partition (where a decoder locates) and then read a bitstream file from external memory, like DDR SDRAM or FLASH. Finally, the circuit would be updated according to the bitstream followed by a ready signal which can be read by ARM cores.

\section{Experimental Preparation and Results}

The whole design is deployed onto a Xilinx ZCU106 MPSoC development kit(Zynq UltraScale+ XCZU7EV-2FFVC1156). All programs concerning image processing and QC-LDPC encoders are supposed to run in the High Performance System (HPS) which is one of major components in this kit, including ARM Cortex A15. The QC-LDPC decoders stay at programming logic (PL, the FPGA part) of ZCU106. Please be noted that encoders are too smaller compared to decoders and therefore, they will not be considered and only realized by software. Both parts are interconnected via AMBA AXI bus and encapsulated in one single chip. Fig. \ref{fig4} is also an experiment platform. It also employs the data collection and scoreboard for comparison.

Two comparison experiments should be conducted in order to select the best QC-LDPC code and analyze those QC-LDPC codes in performance, complexity and power consumption. At last, DPR overhead and strength need to be tested. It is noted that a duplicated design based on \cite{USIT3} participates and there is error-correcting code remained.

\begin{table*}[htbp]
\centering
\caption{The Details of Each Simulated System in the Experiment(1)}
\label{table:Specs}
\begin{tabular}{cccccc}
\toprule  
LDPC Type       &    Without LDPC \cite{USIT3} &  (1920,1280) &  (2016,1680)  & (1920,1600) & (2016,1344) \\
\midrule  

Image Processing Algorithm      & V3 & V3 & V3 & V3 & V3\\
Decoder's Algorithm    & N/A & Min-Sum & Min-Sum&Min-Sum&Min-Sum\\
Decoder's Code Rate    & N/A    & 2/3A    & 2/3A   & 2/3A  & 2/3A  \\
Iteration              & N/A    & 50      & 50     & 50    & 50  \\
\bottomrule 
\end{tabular}
\end{table*}

\begin{table*}[htbp]
\centering
\caption{The Details of Each Simulated System in the Experiment(2)}
\label{table:Specs2}
\begin{tabular}{cccc}
\toprule  
LDPC Type       &    non-V3 (960,640) 2blk &  (960,640) 2blk &  (960,800) 2blk  \\
\midrule  

Image Processing Algorithm      & V1 & V3 & V3\\
Decoder's Algorithm    & Min-Sum & Min-Sum & Min-Sum\\
Decoder's Code Rate    & 2/3A    & 2/3A    & 2/3A   \\
Iteration              & 50     & 50    & 50  \\

\bottomrule 
\end{tabular}
\end{table*}

The iris database used in this paper is CASIA-IrisV4\cite{WB:2002}. We choose a sub-database CASIA-IrisV4-Internal which includes 249 persons' right and left eye images. All eyes are captured for more than 3 times. We create 62,391 inner-comparison groups to get the True Acceptance Rate (TAR). False Acceptance Rate (FAR) derives from 141,913 inter-comparison groups.

In our experiment, we choose performance indexes as below to demonstrate how QC-LDPC codes influence the iris recognition system.

\begin{itemize}
\item FAR: False Acceptance Rate is the measure of the likelihood that the biometric security system will incorrectly accept an access attempt by an unauthorized user.
\item TAR: True Acceptance Rate is the percentage of times a system (correctly) verifies a true claim of identity.
\end{itemize}

\subsection{QC-LDPC Selection}

Considering of the latency of QC-LDPC decoding process and complexity of the proposed system, an ideal class of QC-LDPC codes should fit the size of iris codes generated by the module "Iris Image Processing", which means one block of data includes only one iris image of codes. Two blocks long decoding scheme during one probe phase is also presented for comparison. Compressing 1280 grey-scale pixels, which is the output of iris image processing units, to 1280 black and white pixels is also need to be done. These 1280 pixels, also as known as bits, will be delivered to a QC-LDPC encoder if enrolment phase emergence or a QC-LDPC decoder if probe phase comes.  
\begin{figure}[htbp]
\centerline{\includegraphics[width=0.55\textwidth]{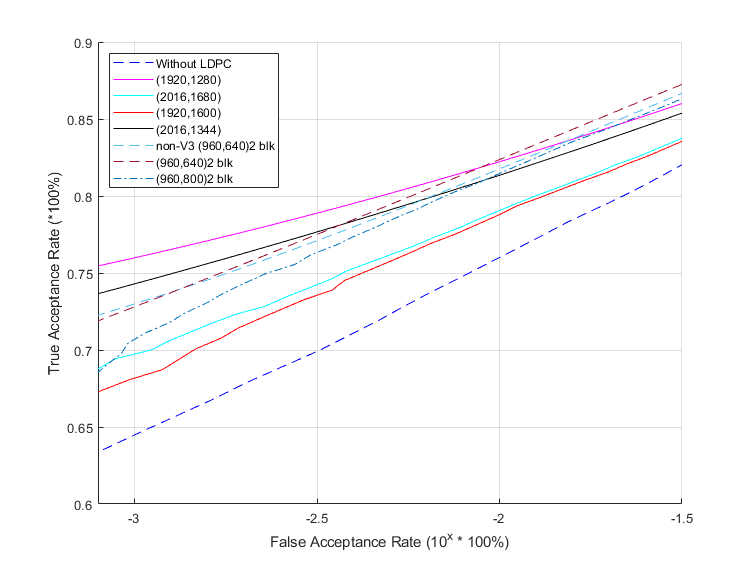}}
    \caption{QC-LDPC codes simulation comparison}
\label{fig10}
\end{figure}
In this experiment, we choose 7 types QC-LDPC codes shown in Table \ref{table:Specs} and Table \ref{table:Specs2}. One type of codes, such as (1920,1280), (2016,1680), (1920,1600) and (2016,1344), are selected because the data to be encoded in these groups are equal or close to 1280 bits which contain iris information and Longer QC-LDPC codes would cause inefficient resource as some dummy ``0" need to be padded.. Another type of codes, like (960,640), are chosen to test whether multiple blocks of encoding and decoding may impact accuracy. Another purpose is to test the parallelism. ``(960,640) 2blk" in Table \ref{table:Specs2} means 1280-bit long input data, if encoded in (960,640), takes two times of operation of this encoder. It takes the same times of operation for the decoder if decoding is needed.   

Fig. \ref{fig10} shows the relationship between FAR and TAR of various objects in the simulation. It can be seen that (1920,1680) outperforms other in most of time, followed by (960,640) 2blk. Increasing the length of QC-LDPC codes, no matter one block or two blocks, doesn't make any improvement. What's worse, The increase in code length causes a performance loss. In terms of image processing algorithm version, the older version, V1 is not as good as V3, comparing non-V3 (960,640) 2blk and (960,640) 2blk.

\subsection{Resource Analysis}

On top of the experiment above, an in-depth study of (1920,1280) and (960,640) 2blk and their implementation should be performed. Table \ref{table:utilization_0} shows the extra Programmable Logic cost, latency and power consumption when a QC-LDPC decoder is implemented. In this table, (960,640) are designed in parallel and serial. The former design includes two decoders operating simultaneously and therefore takes more resources and less processing time. The latter design has to process block by block, take more processing time, but less complex. 

\begin{figure}[htbp]
\centerline{\includegraphics[width=0.55\textwidth]{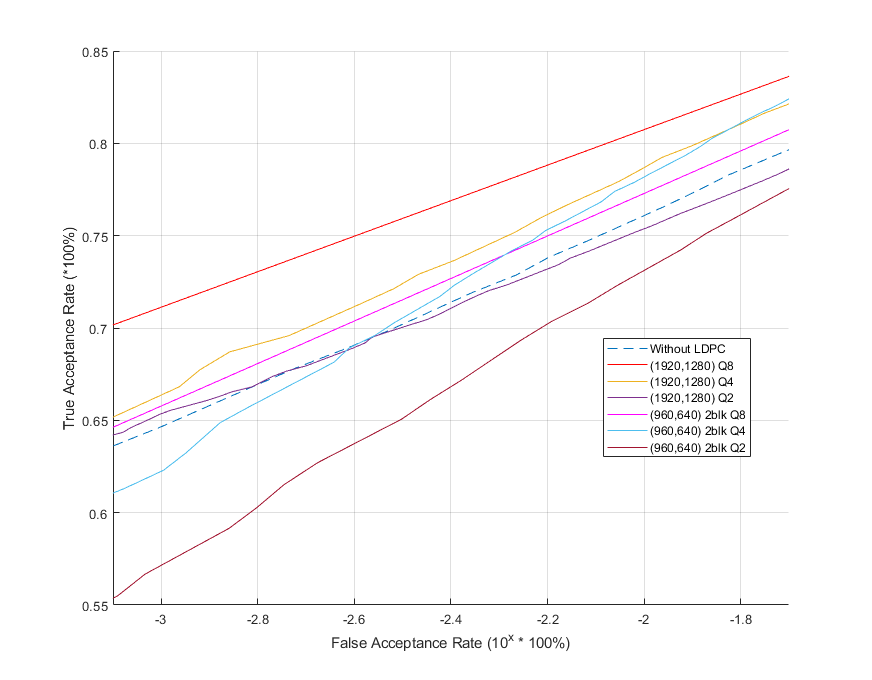}}
    \caption{QC-LDPC codes implementation comparison}
\label{fig20}
\end{figure}

With 8-bit quantization Min-Sum decoding algorithm, Q8 consumes the most hardware resource of all. 18K Block RAM increases linearly as the quantization change from Q2 to Q8 is linear and most large scale of data are stored in 18K Block RAMs. The increase of LUT and FF does not resemble 18K BRAM. They grow slowly since most of control logic is built with them and they are not heavily impacted when quantization bits change. 4-bit quantization seems to be a power-saving option if the accuracy rate result in Fig. \ref{fig20} is considered when some device is power-constraint.  

Note that in Fig. \ref{fig20}, paralleled and serial design for (960,640) decoders have the same accuracy rate, so there is no need to discuss separately.
\begin{table}[htbp]
\centering
\caption{Characters of Synthesized QC-LDPC decoders(1)}
\label{table:utilization_0}
\begin{tabular}{cccc}
\toprule  
 & (1920,1280) & (1920,1280) & (960,640) PARL\\
\midrule  
Q bits       & 8      & 4      & 8      \\
BRAM 18K     & 27     & 14     & 21     \\
FF           & 2422   & 1767   & 4362   \\
LUT          & 10578  & 9500   & 6565   \\
Latency(ns)  & 726839 & 726839 & 368442 \\
Power(mW)    & 35     & 13     & 84     \\
\bottomrule 
\end{tabular}
\end{table}

\begin{table}[htbp]
\centering
\caption{Characters of Synthesized QC-LDPC decoders(2)}
\label{table:utilization_1}
\begin{tabular}{cccc}
\toprule  
 & (960,640) SER & (960,640) PARL & (960,640) SER\\
\midrule  
Q bits       & 8      & 4       & 4\\
BRAM 18K     & 11    & 11      & 5\\
FF           & 2169   & 2895    & 1430 \\
LUT          & 1430   & 4128    & 2032\\
Latency(ns)  & 737518 & 367441  & 735518\\
Power(mW)    & 40     & 38      & 17\\
\bottomrule 
\end{tabular}
\end{table}




\subsection{Dynamic Partial Reconfiguration}

According to the experiment above, we define three different modes, namely fast mode, low-power mode and accuracy mode. Each of the three modes is mapped to a QC-LDPC decoder. Using DPR, the logic resource won't increase linearly as the number of QC-LDPC decoders increases. The QC-LDPC decoders are selected as below:

\begin{itemize}
\item Fast mode: (960,640) in-parallel design with 4-bit quantization spending around 50\% of time compared to others. 
\item Low-Power mode: (1920,1680) with 4-bit quantization only consuming 13mW while others consumes more, sometimes even double or triple quantity.
\item Accuracy mode: (1920,1680) with 8-bit quantization having the best improvement compared to non-LDPC.
\end{itemize}

We implement three QC-LDPC decoders onto one partition of Programmable Logic instead of three partitions. The method is based on \cite{8906146}.In terms of real-time switching, the upload size of QC-LDPC decoder bitstream file is 375KB and it takes 1.75ms to change the whole partition of circuit in order to switch different QC-LDPC decoders.

\section{Conclusions}

In this paper, a multi-QC-LDPC error-correcting iris recognition system is implemented onto n SoC-FPGA platform. Using DPR, resource usage is under reasonable control. Three modes are defined for different scenarios. Such application can meet many requirements of various scenarios, like prompt-respond, high-security, low-power.

With regard to practicability, both image-processing-related modules and LDPC-related modules under our proposed design are implemented in one single chip. We allocate QC-LDPC decoders in a programmable logic section that is much easier for those decoders to be operated in parallel and partially reconfigured. We simulate and find the best QC-LDPC codes and implement them based on the FPGA instead of a traditional program to improve the biometric system running in an embedded device. The whole architecture proves that implementing QC-LDPC codes scheme onto a biometric recognition system is feasible.

\bibliographystyle{IEEEtran}
\bibliography{FPGAGroup}

\end{document}